\documentclass{article}

\PassOptionsToPackage{numbers, sort&compress}{natbib}
\usepackage[dblblindworkshop, preprint]{neurips_2026}
\workshoptitle{World Models for High-Stakes Health (WMHS)}

\usepackage[utf8]{inputenc} 
\usepackage[T1]{fontenc}    
\usepackage{hyperref}       
\usepackage{url}            
\usepackage{booktabs}       
\usepackage{amsfonts}       
\usepackage{amsmath}        
\usepackage{amssymb}
\usepackage{graphicx}
\usepackage{xcolor}
\usepackage{multirow}

\title{Intervention Granularity Matters: Coherent Treatment Bundles in Counterfactual Simulation with Clinical World Models}

\author{%
  Fangzhou Wang\thanks{Equal contribution.} \thanks{Corresponding author: fangzhou.wang@duke.edu}\\
  Duke University \\
  \And
  Yixuan Yang\footnotemark[1] \\
  Duke University \\
  \And
  Camilla Balzarotti \\
  Duke University \\
  \And
  Rishikesan Kamaleswaran \\
  Duke University \\
}

\begin{document}

\maketitle
\vspace{-5mm}

\begin{abstract}
Counterfactual simulation with a clinical world model means fixing a patient's history, changing the treatment, and reading off the predicted response. Doing so requires deciding what counts as one intervention. In clinical settings, interventions are documented as bundles: a co-occurrence audit of 945,707 patient-hours from MIMIC-IV shows groups of components, such as every parameter of a dialysis circuit, that never appear apart, so an edit that changes one component on its own describes an hour that never occurs in the data. We hypothesize that the granularity at which an intervention is edited changes how a world model responds, and test this with Clin-JEPA, a latent world model of patient trajectories conditioned on hourly treatment text. At 1,019 documented onsets of invasive ventilation, we keep the patient's history and other treatments fixed and compare editing one ventilator setting with editing the complete configuration recorded for a real patient with the most similar recent trajectory. The complete bundle moves the predicted next state further than any single setting, consistently across all five settings, and the difference remains after accounting for how much each edit changes the model's input. Intervention granularity therefore materially affects the response of a clinical world model: single-component edits may understate treatment sensitivity, and bundle-aware editing may offer a better-supported basis for counterfactual treatment simulation.
\end{abstract}

\section{Introduction}
\label{sec:intro}

Simulating a treatment arm with a patient world model requires a model that responds to the treatment, and a decision about what counts as one treatment. Action-conditioned world models come from reinforcement learning \citep{ha2018recurrent,hafner2025mastering,assran2025vjepa2}; clinical models now generate patient trajectories from electronic health records \citep{renc2024zero,makarov2025large,kraljevic2024foresight} and sometimes condition the learned transition on interventions \citep{xu2025meddreamer,mu2026ehrworld,wang2026chronomedicalworld,yang2026clinjepa}. Clin-JEPA \citep{yang2026clinjepa} does this for clinical patients: each hour's observations and treatments are written as text, embedded by a language model, and a predictor maps the recent history of state and treatment embeddings to the next state embedding. In principle one can fix a patient's history, change the treatment at one hour, and read off the change in the prediction. The question is what to change.

ICU treatments are not delivered one item at a time. A co-occurrence audit of 945,707 ICU patient-hours (Section~\ref{sec:bundles}) shows groups of components, such as every parameter of a dialysis circuit, or every vasopressor together with the norepinephrine-equivalent dose that summarizes them, that appear only together: the probability of seeing one without the others is zero. An edit that adds or removes such a component on its own therefore describes an hour that never occurs in the training data, a violation of positivity \citep{petersen2012diagnosing,hernan2020causal,gottesman2019guidelines}, and a model that fits the data is free to treat such an input as noise \citep{geirhos2020shortcut}. Action-conditioned world models in other domains are known to disregard actions when the future is predictable from the past \citep{shi2026overcoming,yang2026mirabench}. We therefore hypothesize that the granularity at which an intervention is edited matters: a clinically meaningful treatment may be better represented by the full set of settings that are recorded together than by any one of them alone.

We test this hypothesis on invasive mechanical ventilation, whose bundle (the ventilation status plus the ventilator settings) is explicitly documented. We propose bundle-consistent editing: at a documented onset of ventilation, the ventilation description is replaced by the complete configuration recorded for the real patient with the most similar 24-hour trajectory (Section~\ref{sec:bundle}). On 1,019 onsets, the complete bundle produces a larger next-hour response than any single setting, and the difference remains after adjusting for how far each edit moves the treatment embedding (Section~\ref{sec:results}).

\begin{figure}[t]
\centering
\includegraphics[width=0.86\linewidth]{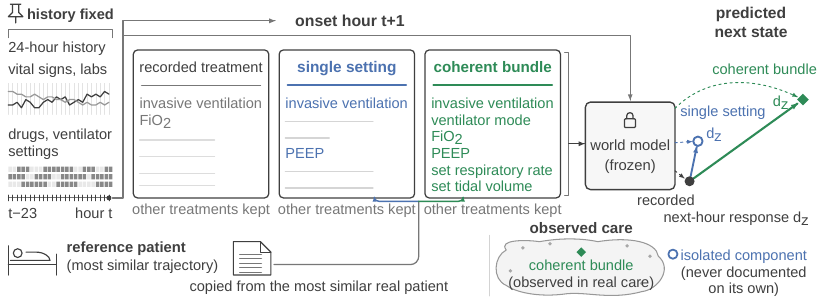}
\caption{\textbf{Bundle-consistent editing at a ventilation onset.} The 24-hour history is fixed. At the onset hour, the ventilation description is replaced by one setting or by the complete configuration recorded for the reference patient, the real patient with the most similar trajectory; other treatments are kept. The frozen world model predicts the next state under each text, and the response $d_z$ is the relative distance from the prediction under the recorded text. Inset: a coherent bundle is a configuration observed in real care; a component on its own is not.}
\label{fig:overview}
\end{figure}
\vspace{-5mm}
\section{Interventions in ICU records come as bundles}
\label{sec:bundles}

In the data used by Clin-JEPA \citep{yang2026clinjepa,johnson2023mimic}, every ICU hour has a treatment text that lists that hour's interventions (drugs with doses and rates, ventilator and dialysis settings, fluids, procedures); hours without interventions receive the sentence ``No active interventions.'' We audited how the components of these texts co-occur in the test split (945,707 unique patient-hours from 12,631 stays). For six common ICU interventions we counted every component present in the same hour as the intervention and computed its \emph{co-occurrence ratio} $\rho(B\mid I)=P(B\mid I)/P(B)$: how many times more often component $B$ appears in hours with intervention $I$ than in hours overall (Appendix~\ref{app:cooccurrence}). For blood-pressure management (146,829 hours with an arterial line or a vasopressor), the seven vasoactive agents and the norepinephrine-equivalent dose, a summary of all vasopressors in one unit \citep{kotani2023updated}, all have exactly the same ratio, $6.44=945{,}707/146{,}829$; for dialysis (40,604 hours), twenty parameters of the renal-replacement circuit all have exactly the same ratio, $23.29=945{,}707/40{,}604$. A ratio equal to $N/N_I$ ($N$ hours in total, $N_I$ with the intervention) arises only when every hour containing the component contains the intervention, $P(B\mid\neg I)=0$: these components are never documented apart from the intervention they belong to, and the ventilator settings used in Section~\ref{sec:bundle} behave the same way inside the ventilation bundle. This is how protocol-driven ICU care \citep{jaber2010intervention,russotto2021intubation,klompas2022strategies} appears in the record: as bundles of co-interventions.

Two design rules for counterfactual edits follow. First, an edit that adds a ventilator setting to an unventilated patient, changes one dialysis parameter on its own, or removes one vasopressor while leaving the summary dose unchanged describes an hour with zero support in the data; components that only occur together should be edited together. Second, derived summaries such as the norepinephrine-equivalent dose (present in 88.3\% of blood-pressure-management hours and never outside them) are computed from the individual doses and must be recomputed after an edit; changing a dose without updating the summary produces a text that contradicts itself.

\section{Bundle-consistent editing at invasive ventilation onset}
\label{sec:bundle}

\paragraph{Model and measurements.}
We use the released Clin-JEPA checkpoint \citep{yang2026clinjepa} without retraining. Each hour $t$ of a stay is written as a state text $s_t$ (vital signs, laboratory results, scores) and the treatment text $a_t$ described above. A Qwen3-8B encoder with LoRA adapters maps each text to a 4096-dimensional embedding, $z_t=E(s_t)$ and $u_t=E(a_t)$, and a 92M-parameter transformer predictor reads the interleaved embeddings of the previous 24 hours and predicts the next state embedding $\hat z_{t+1}$. All quantities below compare the model's own predictions; no observed future serves as counterfactual ground truth. When the history is fixed and one hour's treatment text is edited, the \emph{next-hour response} is the relative distance between the prediction under the edited text and the prediction under the recorded text, $d_z=\lVert\hat z^{\mathrm{edit}}-\hat z^{\mathrm{rec}}\rVert_2/\lVert\hat z^{\mathrm{rec}}\rVert_2$, and the \emph{edit magnitude} is the distance between the two treatment embeddings, $d_a=\lVert u^{\mathrm{edit}}-u^{\mathrm{rec}}\rVert_2$.

\paragraph{Bundle and configuration bank.}
We test the granularity hypothesis on one intervention with a clear bundle, invasive mechanical ventilation, at its onset. The index component is the ventilation status \texttt{InvasiveVent}; the clinician-set components are ventilator mode, FiO$_2$, PEEP, set respiratory rate, and set tidal volume. Measured consequences of ventilation, such as plateau pressure and minute volume, are excluded: they are readouts, not decisions. From the test split we built a bank of every invasively ventilated hour (229,037 hours, 4,031 stays), keeping for each hour the subset of the five settings actually documented (66.4\% record only the status, 16.5\% all five).

\paragraph{Decision points.}
A target is an hour $t$ without invasive ventilation followed by an hour $t+1$ with it, with no invasive ventilation in the preceding 12 hours, a complete 24-hour history, at least 12 further hours of the stay, and ventilation persisting for at least two hours. Of 3,553 raw transitions in the test split, 1,089 (788 stays) qualify; only 1.5\% had six or more hours of non-invasive ventilation in the preceding day, so late rescue after failed non-invasive support is rare (Appendix~\ref{app:cohort}).

\paragraph{Reference bundles from trajectory-matched patients.}
Rather than writing ventilator settings by hand, we take them from a real patient: for each target we compared its 24 state embeddings $z_{t-23},\ldots,z_t$ with the 24 pre-onset state embeddings of every other documented onset in the test split (1,130 eligible onsets), using the mean of the hourly cosine similarities, and took the most similar onset from a different stay as the reference; treatment embeddings and post-onset states play no role. Every target found a reference (mean similarity 0.929), 88.2\% of reference bundles contain at least three settings and 52.6\% all five, and masking the six hours before onset from the matching changes none of this (Appendix~\ref{app:matching}).

\paragraph{Conditions and endpoint.}
For every target we build three versions of the onset-hour treatment text, keeping every non-ventilation fragment as recorded and changing only the ventilation description (Figure~\ref{fig:overview}): the \emph{recorded} text; the \emph{single-setting} text, which keeps the ventilation status and copies one setting from the reference bundle, one condition per setting the reference documents; and the \emph{bundle} text, which keeps the status and copies every setting the reference documents. Each version is re-encoded and passed to the frozen predictor with the unchanged history, and the endpoint is the next-hour response $d_z$. We compare the bundle with each single-setting condition on the same targets by the paired difference in $d_z$ with a bootstrap 95\% confidence interval (5,000 resamples). To check that the difference is not a matter of how far each edit moves the treatment embedding, we also fit the linear model $d_z=\beta_0+\beta_b\,\mathbf{1}[\text{bundle}]+\beta_1 d_a+\varepsilon$ on the single-setting and bundle rows of the same targets (with a $d_a^2$ term as a sensitivity analysis); the \emph{adjusted} bundle effect is $\beta_b$, the extra response attributed to the bundle at equal edit magnitude. We stop at the next hour on purpose: rolling further would require deciding what treatments follow under each condition.

\section{Results}
\label{sec:results}

\begin{figure}[t]
\centering
\includegraphics[width=\linewidth]{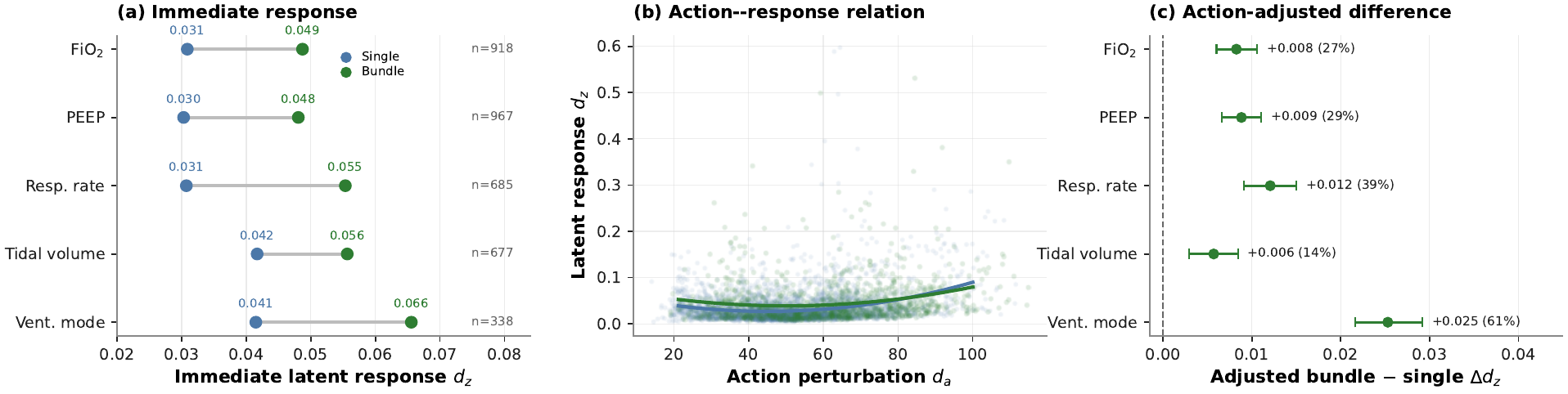}
\caption{
\textbf{Immediate representation responses to ventilation intervention construction.}
\textbf{(a)} Mean factual-relative next-state latent response $d_z$ for single-component (blue) and coherent-bundle (green) edits on paired targets; $n$ is the number of paired target instances for each setting.
\textbf{(b)} Relationship between action-embedding perturbation magnitude $d_a$ and $d_z$; curves are descriptive quadratic fits (overall Pearson $r=0.21$).
\textbf{(c)} Action-distance-adjusted bundle-minus-single differences in $d_z$ with 95\% target-level bootstrap confidence intervals. Percentages give the adjusted difference relative to the corresponding mean single-component $d_z$. (Appendix~\ref{app:scatter})
}
\label{fig:results}
\end{figure}

1,019 of the 1,089 targets have a usable reference bundle and enter the comparison; the remaining 70 matched reference bundles contain no eligible clinician-controlled ventilator setting from which to construct a bundle edit. Table~\ref{tab:main} and Figure~\ref{fig:results} give the mean next-hour response under each condition. On the same patients, the bundle moves the prediction further from the prediction under the recorded text than any single ventilator setting: the paired difference is 0.018 for FiO$_2$, 0.018 for PEEP, 0.025 for set respiratory rate, 0.014 for set tidal volume, and 0.024 for ventilator mode, and every bootstrap confidence interval lies above zero. Adjusting for the edit magnitude changes the picture little (Appendix~\ref{app:scatter}): at equal edit magnitude the bundle advantage remains positive for each setting, between 0.006 and 0.025, which is 14\% to 61\% of the corresponding single-setting response, and the pooled effect is $+0.0096$ with a linear and $+0.0080$ (95\% CI 0.005 to 0.011) with a quadratic adjustment. The bundle produces the larger response in 70.6--89.5\% of paired targets across settings. Mean $d_a$ ranges from 45.1 to 68.3 across intervention constructions and is not uniformly larger for bundle edits. Excluding the six hours immediately preceding onset also leaves the reference-matching neighborhood largely stable, with 98.2\% of masked-window Top-1 references remaining within the original Top-20 neighborhood. The larger immediate response to bundle edits is therefore not fully explained by action-embedding perturbation magnitude alone.

\begin{table}[t]
\centering
\scriptsize
\setlength{\tabcolsep}{3pt}
\caption{\textbf{The complete bundle produces a larger next-hour response than any single setting.} Mean next-hour response $d_z$ under the single-setting and the bundle condition, on the targets whose reference documents the setting in the row. Bundle $-$ single: mean paired difference with its bootstrap 95\% confidence interval. Adjusted: the bundle effect $\beta_b$ at equal edit magnitude. Adj.\,/\,single: the adjusted effect as a fraction of the single-setting mean, for scale only.}
\label{tab:main}
\begin{tabular}{lrcclcc}
\toprule
Setting & $n$ & Single setting & Coherent bundle & Bundle $-$ single [95\% CI] & Adjusted & Adj.\,/\,single \\
\midrule
FiO$_2$              & 918 & 0.0309 & 0.0487 & $+0.0178$ [0.0157, 0.0201] & $+0.0083$ & 27\% \\
PEEP                 & 967 & 0.0303 & 0.0481 & $+0.0178$ [0.0157, 0.0199] & $+0.0088$ & 29\% \\
Set respiratory rate & 685 & 0.0307 & 0.0553 & $+0.0246$ [0.0220, 0.0274] & $+0.0121$ & 39\% \\
Set tidal volume     & 677 & 0.0417 & 0.0556 & $+0.0140$ [0.0112, 0.0167] & $+0.0057$ & 14\% \\
Ventilator mode      & 338 & 0.0415 & 0.0656 & $+0.0241$ [0.0202, 0.0281] & $+0.0253$ & 61\% \\
\bottomrule
\end{tabular}
\end{table}

\section{Discussion and limitations}
\label{sec:discussion}

Intervention granularity changes what a clinical world model predicts. Under identical histories, a coherent configuration taken from a similar real patient produces a clearly larger immediate response than any of its components alone, and the gap is not a matter of edit size. With the audit, this suggests how to construct counterfactual treatment arms: define the unit of intervention as a bundle found in the data and reviewed clinically; take its values from observed care for similar patients, and report the reference similarity with the effect; and recompute derived summary variables after every edit. Target-trial emulation asks the same of observational analyses, a well-defined and sustained intervention \citep{hernan2016using}, and counterfactual models over time estimate treatment regimes rather than single actions \citep{lim2018forecasting,bica2020estimating,li2021gnet}. For treatment-arm simulation \citep{thorlund2020synthetic}, single-component edits may understate a model's treatment sensitivity, and bundle-aware editing may offer a better-supported basis.

The claims are limited to representation: a larger response shows that the model distinguishes the bundle from its components, not that the unobserved counterfactual is correct. Because single-setting configurations also occur in the data, this experiment tests edit granularity rather than empirical support; unsupported edits require direct evaluation. 

\clearpage
\bibliographystyle{plainnat}
\bibliography{references}

\clearpage
\section*{Responsible-use statement}
This work uses the de-identified MIMIC-IV database under its data use agreement. The world model and the editing protocol are research tools for evaluating models; neither is validated for clinical use, and no output is a basis for an individual treatment decision. One implication is cautionary: a model that appears to support counterfactual simulation may respond to interventions for reasons unrelated to their clinical effect, and the effect sizes read from it depend on how the intervention is specified. We make no claim about the benefit or harm of intubation. Any use of simulated treatment arms as evidence should report the data support of the edited configurations, the similarity of the reference patients, and the sensitivity of the conclusion to the choice of bundle. Responses may also differ across patient groups unevenly represented in ICU data, which we have not examined.

\clearpage
\appendix

\section{Decision-point cohort}
\label{app:cohort}

\begin{table}[h]
\centering
\small
\caption{Cohort flow for invasive-ventilation onset targets in the test split. The primary analysis uses the 12-hour washout (bold).}
\label{tab:cohort}
\begin{tabular}{lrr}
\toprule
Criterion & Target hours & Stays \\
\midrule
Raw transition (no InvasiveVent at $t$, InvasiveVent at $t+1$) & 3,553 & 2,903 \\
\addlinespace
6-h washout & 3,442 & 2,877 \\
\quad + 24-h history + 12-h future & 1,127 & 805 \\
\quad + persistent $\geq$ 2 h & 1,125 & 803 \\
\addlinespace
12-h washout & 3,404 & 2,869 \\
\quad + 24-h history & 1,125 & 809 \\
\quad + 12-h future & 1,091 & 790 \\
\quad \textbf{+ persistent $\geq$ 2 h (primary cohort)} & \textbf{1,089} & \textbf{788} \\
\addlinespace
24-h washout & 3,108 & 2,813 \\
\quad + 24-h history + 12-h future + persistent $\geq$ 2 h & 816 & 675 \\
\bottomrule
\end{tabular}
\end{table}

\paragraph{Respiratory support before onset.} Late intubation after prolonged non-invasive support is a clinically distinct path with worse outcomes \citep{kangelaris2016timing}, so we checked how common it is among the 1,089 targets. Non-invasive ventilation in the preceding 24 hours: 0 h for 1,069 targets (98.2\%), under 6 h for 4, 6 to 12 h for 6, and 12 to 24 h for 10. The documented ventilation status at hour $t$ is: none, 749 (68.8\%); supplemental oxygen, 24.5\%; high-flow nasal cannula, 4.0\%; tracheostomy, 1.7\%; non-invasive ventilation, 1.0\%. Of the 749 targets with no documented status at $t$, 449 have no documented status anywhere in the preceding 24 hours; for the other 300 the last documented status was invasive ventilation (225, all 12 to 24 hours before $t$, consistent with re-initiation rather than a first decision), supplemental oxygen (61), tracheostomy (6), high-flow nasal cannula (5), or non-invasive ventilation (3). SpO$_2$ is available at $t$ for 87.2\% of the 749 but an oxygen flow rate for only 2.3\%, so we did not impute a status for this heterogeneous group. The targets occur a median of 95 hours after ICU admission (interquartile range 49 to 171, range 23 to 323).

\section{Reference matching and masked-window sensitivity}
\label{app:matching}

To reduce sensitivity to documentation immediately preceding ventilation onset, masked-window matching excludes the final six pre-onset hours from the similarity calculation, using only state embeddings from $t-23$ through $t-6$; the target onset time and subsequent intervention construction are unchanged.

\begin{table}[h]
\centering
\footnotesize
\caption{Reference matching with the full 24-hour window versus the window with the last six hours before onset masked. Similarity: mean hourly cosine similarity of the best match.}
\label{tab:matching}
\begin{tabular}{lcc}
\toprule
 & Full window ($t-23$ to $t$) & Masked window ($t-23$ to $t-6$) \\
\midrule
Targets matched & 1,089 / 1,089 & 1,089 / 1,089 \\
Mean similarity of best match & 0.929 & 0.934 \\
5th percentile of similarity & 0.894 & 0.899 \\
Distinct references selected & 518 & 541 \\
Reuse per reference: median / 95th pct.\ / max & 1 / 5 / 12 & 1 / 5 / 15 \\
Bundles with $\geq 3$ of 5 settings & 88.2\% & 88.3\% \\
Bundles with all 5 settings & 52.6\% & 55.2\% \\
Same best match under both windows & \multicolumn{2}{c}{59.2\%} \\
Masked best match within full-window top 20 & \multicolumn{2}{c}{98.2\%} \\
\bottomrule
\end{tabular}
\end{table}

\section{Co-occurrence audit}
\label{app:cooccurrence}

The audit uses one row per unique (stay, hour) pair of the test split ($N=945{,}707$) and treats every fragment of the treatment text as a component (repeated mentions in one hour count once). The six interventions are invasive ventilation, blood-pressure management, oxygen escalation, dialysis, neuromuscular blockade, and antipsychotic treatment. Intervention hours are defined from explicit components: invasive ventilation from the status field; blood-pressure management from an arterial-line record or any explicit vasoactive agent, with the derived norepinephrine-equivalent dose deliberately excluded from the definition; dialysis from the renal-replacement status; and likewise for the other three. For each intervention, components are ranked by the co-occurrence ratio after support gates (at least 10 co-occurring hours, at least 20 hours overall, presence in at least 0.5\% of intervention hours). Identical ratios within a group identify components that occur only inside the intervention. For blood-pressure management, eight components had the same ratio of 6.44, including the seven intervention-defining vasoactive agents and the norepinephrine-equivalent dose. For dialysis, twenty dialysis/CRRT components were similarly fully contained within dialysis hours (ratio 23.29). Notably, although the norepinephrine-equivalent dose was excluded from the blood-pressure-management definition, it appeared in 88.3\% of blood-pressure-management hours and never outside them.

\section{Response versus edit magnitude}
\label{app:scatter}

This figure plots the next-hour response $d_z$ against edit magnitude $d_a$ for every single-setting (blue) and bundle (green) edit. The response rises only weakly with $d_a$ (Pearson correlation 0.21), but the  main comparison still adjusts for it for the sake of fairness; the trend lines are descriptive.

\end{document}